\documentclass[10pt,twocolumn,letterpaper]{article}

\usepackage{iccv}
\usepackage{times}
\usepackage{epsfig}
\usepackage{graphicx}
\usepackage{amsmath}
\usepackage{amssymb}
\usepackage{multirow}
\usepackage{lipsum}
\usepackage{adjustbox}
\usepackage{booktabs}
\usepackage{makecell}
\usepackage{tabu}
\usepackage{bm}
\usepackage{xcolor}

\DeclareMathOperator{\CLIPTextEncoder}{CLIPTextEncoder}
\DeclareMathOperator{\MLP}{MLP}
\newcommand{\epstheta}{$\bm{\epsilon}_\theta$}
\newcommand{\paragrapha}[2][4pt]{\vspace{#1}\noindent\textbf{#2}}

\usepackage[pagebackref=true,breaklinks=true,letterpaper=true,colorlinks,bookmarks=false]{hyperref}

\usepackage{cleveref}
\iccvfinalcopy 

\def\iccvPaperID{3043} 

\ificcvfinal\fi

\usepackage{pifont}
\newcommand{\cmark}{\ding{51}}%
\newcommand{\xmark}{\color{gray!30!white}\ding{55}}%

\begin{document}


\title{Unleashing Text-to-Image Diffusion Models for Visual Perception}

\author{
\and
Second Author\\
Institution2\\
First line of institution2 address\\
{\tt\small secondauthor@i2.org}
}


\def\spaces{~~~~~~}
\author{Wenliang Zhao\textsuperscript{1}\thanks{Equal contribution. ~~\textsuperscript{\dag}Corresponding authors.}\spaces{}Yongming Rao\textsuperscript{1}\footnotemark[1]\spaces{}Zuyan Liu\textsuperscript{1}\footnotemark[1]\spaces{}Benlin Liu\textsuperscript{2}\spaces{}Jie Zhou\textsuperscript{1}\spaces{}Jiwen Lu$^{1\dagger}$\\\\
\textsuperscript{1}Tsinghua University~~
\textsuperscript{2}University of Washington}

\maketitle
\ificcvfinal\thispagestyle{empty}\fi

\begin{abstract}
Diffusion models (DMs) have become the new trend of generative models and have demonstrated a powerful ability of conditional synthesis. Among those, text-to-image diffusion models pre-trained on large-scale image-text pairs are highly controllable by customizable prompts. Unlike the unconditional generative models that focus on low-level attributes and details, text-to-image diffusion models contain more high-level knowledge thanks to the vision-language pre-training. In this paper, we propose VPD (Visual Perception with a pre-trained Diffusion model), a new framework that exploits the semantic information of a pre-trained text-to-image diffusion model in visual perception tasks. Instead of using the pre-trained denoising autoencoder in a diffusion-based pipeline, we simply use it as a backbone and aim to study how to take full advantage of the learned knowledge. Specifically, we prompt the denoising decoder with proper textual inputs and refine the text features with an adapter, leading to a better alignment to the pre-trained stage and making the visual contents interact with the text prompts. We also propose to utilize the cross-attention maps between the visual features and the text features to provide explicit guidance. Compared with other pre-training methods, we show that vision-language pre-trained diffusion models can be faster adapted to downstream visual perception tasks using the proposed VPD. Extensive experiments on semantic segmentation, referring image segmentation and depth estimation demonstrates the effectiveness of our method. Notably, VPD attains 0.254 RMSE on NYUv2 depth estimation  and 73.3\% oIoU on RefCOCO-val referring image segmentation, establishing new records on these two benchmarks. Code is available at \url{https://github.com/wl-zhao/VPD}.

\end{abstract}

\section{Introduction}

Recently, large text-to-image diffusion models~\cite{rombach2022high,ramesh2022dalle2} have demonstrated phenomenal power in generating diverse and high-fidelity images with high customizability~\cite{rombach2022high,hertz2022prompt,parmar2023zero,brooks2022instructpix2pix}, attracting growing attention from both the research community and the public eye. By leveraging large-scale datasets of image-text pairs (\eg, LAION-5B~\cite{schuhmann2022laion}), text-to-image diffusion models exhibit favorable scaling ability. Large-scale text-to-image diffusion models are able to generate high-quality images with rich texture, diverse content and reasonable structures while having compositional and editable semantics. This phenomenon potentially suggests that large text-to-image diffusion models can \emph{implicitly} learn both high-level and low-level visual concepts from massive image-text pairs. Moreover, recent research~\cite{hertz2022prompt,parmar2023zero} also has highlighted the clear correlations between the latent visual features and corresponding words in text prompts in text-to-image diffusion models.

\begin{figure}[t]
    \centering
    \includegraphics[width=\linewidth]{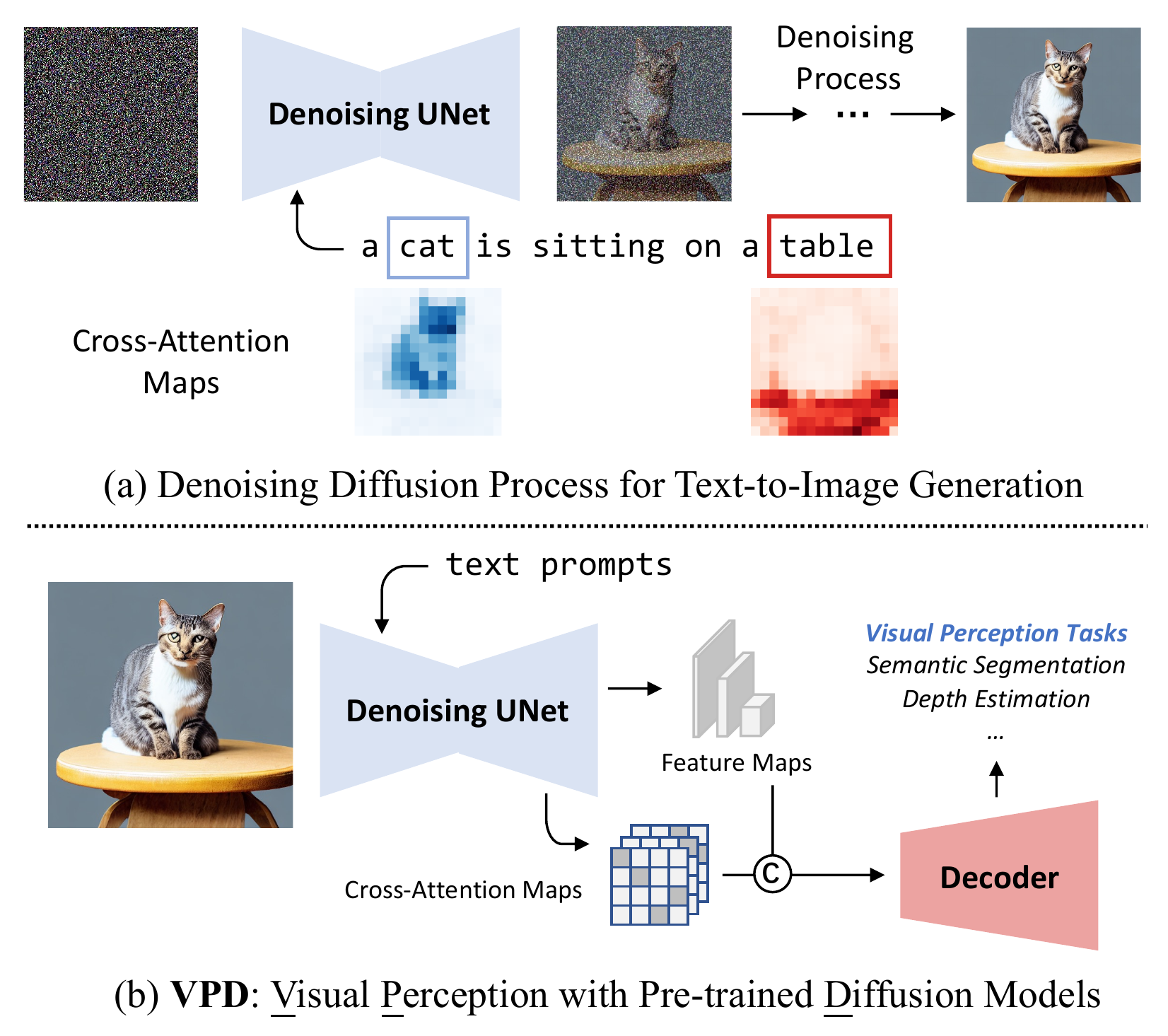}
    \caption{\textbf{The main idea of the proposed VPD framework.} Motivated by the compelling generative semantic of a text-to-image diffusion model, we proposed a new framework named VPD to exploit the pre-trained knowledge in the denoising UNet to provide semantic guidance for downstream visual perception tasks.}
    \label{fig:idea}
    \vspace{-15pt}
\end{figure}

The compelling generative semantic and compositional abilities of text-to-image diffusion models motivate us to think: \emph{is it possible to extract the visual knowledge learned by large diffusion models for visual perception tasks? } However, it is non-trivial to solve this problem. Conventional visual pre-training methods aim to encode the input image as latent representations  and learn the representations with pretext tasks like contrastive learning~\cite{he2019moco,chen2020simple} and masked image modeling~\cite{bao2021beit,he2022mae} or massive annotations in classification and vision-language tasks. The pre-training process makes the learned latent representation naturally suitable for a range of visual perception tasks as semantic knowledge is extracted from the raw images. In contrast, text-to-image models are designed to generate high-fidelity images based on textual prompts.  Text-to-image diffusion models take as input random noises and text prompts, and aim to produce images through a progressive denoising process~\cite{rombach2022high,ho2020ddpm}.  While there is a notable gap between the text-to-image  generation task and the conventional visual pre-training mechanisms, the training process of text-to-image models also requires them to capture both low-level knowledge of images (\eg, textures, edge, and structures) and high-level semantic relations between visual and linguistic concepts from diverse and large-scale image-text pairs in an implicit way. Although rich representations are learned in large diffusion models, it is still unknown how to extract this knowledge for various visual perception tasks and whether it can benefit visual perception.

In this paper, we study how to leverage the knowledge learned in text-to-image for visual perception. Compared to transferring knowledge from conventional pre-trained models to downstream visual perception tasks, there are two distinct challenges to performing transfer learning on diffusion models:  the incompatibility between the diffusion pipeline and visual perception tasks and the architectural differences between UNet~\cite{ronneberger2015unet}-like diffusion models and popular visual backbones. To tackle these challenges, we introduce a new framework called \emph{VPD} to adapt pre-trained diffusion models for visual perception tasks.  Instead of using the step-by-step diffusion pipeline, we propose to simply employ the autoencoder as a backbone model to directly consume the natural images without noise and perform a single extra denoising step with designed prompts to extract the semantic information. Our framework is based on popular Stable Diffusion~\cite{rombach2022high} models, which conduct the denoising process in a learned latent space with a UNet architecture.  We  extract features from different hierarchies from the UNet decoder to construct visual representations of the input image. To align with the pre-trained stage and facilitate interactions between visual content and text prompts, we  prompt the denoising diffusion model with proper textual inputs and refine the text features with an adapter. Additionally, inspired by previous studies on the relations between prompt words and visual patterns in diffusion models, we propose to utilize the cross-attention maps between the visual and text features to provide explicit guidance. The combined implicit and explicit guidance can be fed to various visual decoders to perform visual perception tasks.   Our main idea is summarized in Figure~\ref{fig:idea}.

We evaluate our method on three representative visual perception tasks covering: 1) semantic segmentation~\cite{zhou2017ade} which requires the understanding of high-level and fine-grained visual concepts, 2) referring image segmentation~\cite{yu2016refcoco,nagaraja2016gref_umd} that requires the ability of visual-language modeling, and 3) depth estimation~\cite{silberman2012nyuv2} that requires low-level and structural knowledge of images. With the help of the proposed VPD,  we show that a vision-language pre-trained diffusion model can be a fast and powerful learner of downstream visual perception tasks. Our method attains 73.3\% oIoU and 0.254 RMSE on RefCOCO~\cite{yu2016refcoco} referring image segmentation and NYUv2~\cite{silberman2012nyuv2} depth estimation, respectively, establishing new state-of-the-art on these two benchmarks. Equipped with a lightweight Semantic FPN~\cite{kirillov2019semanticfpn} decoder, our model achieves 54.6\% mIoU on ADE20K~\cite{zhou2017ade}, outperforming supervisedly pre-trained ConvNeXt-XL~\cite{liu2022convnet} model with comparable computational complexity. We also exhibit that models pre-trained with diffusion tasks can fast obtain 44.7\% mIoU on this challenging benchmark with only 4K iteration training, outperforming existing pre-training methods. We expect our study to offer a new perspective on learning more generic visual representations with generative models and spark further research on bridging and unifying the vibrant research fields of image generation and perception. 

\begin{figure*}[t]
\begin{center}
\includegraphics[width=0.9\linewidth]{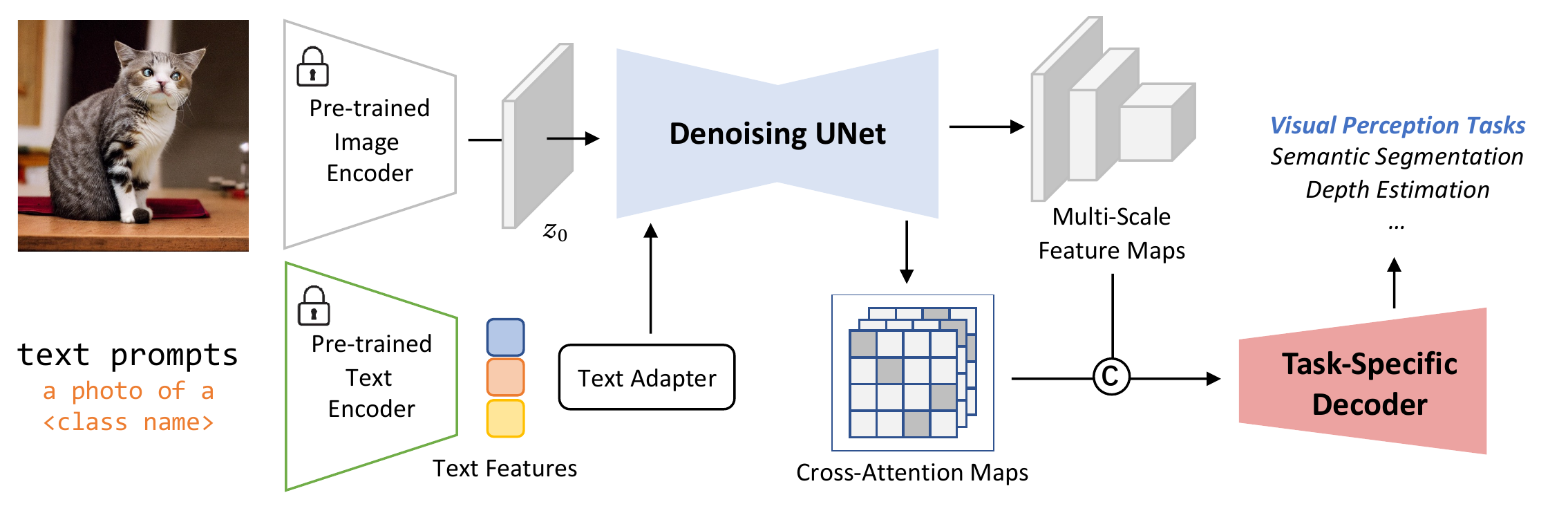}
\end{center}
   \caption{\textbf{The overall framework of VPD.} To better exploit the semantic knowledge learned from text-to-image generation pre-training, we prompt the denoising UNet with properly designed text prompts and employ the cross-attention maps to provide both implicit and explicit guidance to downstream visual perception tasks. Our framework can fully leverage both the low-level and high-level pre-trained knowledge and can be applied in a variety of visual perception tasks.}
\label{fig:overall}
\vspace{-10pt}
\end{figure*}

\section{Related Work}
\paragrapha{Diffusion Models.} Diffusion denoising probabilistic models, also known as diffusion models, have emerged as a new prevailing family of generative models that demonstrate remarkable synthesis quality and controllability. The fundamental concept behind the diffusion model involves training a denoising autoencoder to learn the inverse of a Markovian diffusion process~\cite{sohl2015deep,ho2020ddpm}. With proper re-parameterization, the training objective of diffusion models can be formulated as a simple weighted MSE loss~\cite{ho2020ddpm}, which makes diffusion models enjoy more stable training compared with GANs~\cite{goodfellow2014generative} and VAEs~\cite{kingma2013auto}. Sampling from a diffusion model~\cite{song2020denoising,liu2022pseudo,lu2022dpmsolver} can then be viewed as a progressive denoising procedure, which requires multiple evaluations of the denoising autoencoder. 
As a step towards high-resolution image synthesis based on diffusion models, Rombach~\etal~\cite{rombach2022high} propose the latent diffusion models (LDMs), which perform diffusion on a latent space of a lower resolution and thus can significantly reduce the computational costs. They also propose a generic solution to add conditions via the cross-attention~\cite{vaswani2017attention} mechanism. These advancements allow for training text-to-image diffusion models on a large-scale dataset LAION-5B~\cite{schuhmann2022laion}, which are now available in the famous ``Stable-Diffusion'' library. Recent work by~\cite{hertz2022prompt} has witnessed a clear visual-text correlation in the large text-to-image diffusion models, which motivates us to study whether the pre-trained knowledge can be exploited to facilitate downstream visual perception tasks. Different from previous diffusion-based framework~\cite{chen2022diffusiondet,amit2021segdiff} that reformulate the visual perception task as progressive denoising, we employ the denoising autoencoder pre-trained on the text-to-image generation as a backbone and study how to make full use of the learned high-level and low-level knowledge, which only require a single forward pass of the denoising autoencoder.


\paragrapha{Visual Pre-training.} The pre-training \& fine-tuning paradigm has significantly pushed the development of computer vision, especially in downstream visual perception tasks where labels are hard to collect. The most widely used pre-training is supervised pre-training on large-scale image classification datasets like ImageNet~\cite{deng2009imagenet}. Besides, self-supervised learning such as contrastive learning~\cite{caron2021dino,he2019moco} and masked image modelling~\cite{peng2022beitv2,he2022mae} have also proved to be able to learn transferrable representations. In this paper, we will demonstrate that large-scale text-to-image generation can also be a possible alternative for visual pre-training. Different from the standard visual pre-training methods that are specifically designed for extracting high-level representation of visual data, a model trained on a generative task focuses on the synthesis quality and captures more low-level clues. However, our results show that due to the existence of natural language during pre-training, a well-learned text-to-image diffusion model contains sufficient both high-level and low-level knowledge, which can also be applied in downstream visual perception tasks.

\section{Method}\label{sec:method}
In this section, we present VPD, a new framework that achieves visual perception with a pre-trained diffusion model. Our key idea is to investigate how to fully extract the pre-trained high-level knowledge in a pre-trained text-to-image diffusion model. We will start by reviewing the background of diffusion models, and then describe our designs of VPD, including how to implicitly and explicitly leverage the visual-language correspondence lies in the pre-trained text-to-image diffusion models. The overall framework of our VPD is illustrated in \Cref{fig:overall}.

\subsection{Preliminaries: Diffusion Models}

To begin with, we will provide a brief overview of the diffusion models~\cite{sohl2015deep, ho2020ddpm, kingma2021variational}. Diffusion models are a new family of generative models that can reconstruct the distribution of data by learning the reverse process of a diffusion process. Denoting $\bm{z}_{t}$ as the random variable at $t$-th timestep, the diffusion process is modeled as a Markov: 
\begin{equation}
    \bm{z}_{t} \sim \mathcal{N}(\sqrt{\alpha_t}\bm{z}_{t-1}, (1-\alpha_t) \bm{I}),
\end{equation}
where $\{\alpha_t\}$ are fixed coefficients that determine the noise schedule. The above definition leads to a simple close form of $p(\bm{z}_t|\bm{z}_0)$:
\begin{equation}
\begin{split}
    &\bm{z}_t=\sqrt{\bar{\alpha}_t} \bm{z}_0 +  \sqrt{1-\bar{\alpha}_t}\bm{\epsilon}\\
    &\bar{\alpha}_t=\prod_{s=1}^t\alpha_s, \bm{\epsilon}\sim\mathcal{N}(\bm{0}, \bm{I}),
\end{split}
    \label{equ:repzt}
\end{equation}
which further allows sampling an arbitrary $\bm{z}_t$ efficiently during training. With proper re-parameterization, the training objective of diffusion models can be derived as~\cite{ho2020ddpm}:
\begin{equation}
    L_{\rm DM} = \mathbb{E}_{\bm{z}_0, \bm{\epsilon}, t}\left[\|\bm{\epsilon} - \bm{\epsilon}_\theta(\bm{z}_t(\bm{z}_0, \bm{\epsilon}), t; \mathcal{C})\|^2_2\right],
    \label{equ:simple}
\end{equation}
where $\bm{z}_t$ is computed as Equation~\eqref{equ:repzt}. $\bm{\epsilon}_\theta$ is an autoencoder (usually implemented as a UNet~\cite{ronneberger2015unet}) that is learned to predict the $\bm{\epsilon}$ given the conditioning inputs $\mathcal{C}$. The sampling of diffusion models is achieved by discretizing the diffusion SDE or ODE~\cite{song2021score} thus requires multiple model evaluations at different timesteps.

The training objective~\eqref{equ:simple} enables stable training of diffusion models, even with complex conditioning inputs. Recently, \cite{rombach2022high} released a text-to-image model (namely ``Stable-Diffusion'') trained on large-scale image-text dataset LAION-5B~\cite{schuhmann2022laion}, which has demonstrated remarkable performance on image synthesis controlled by natural language. Specifically, they first train a VQGAN consisting of an encoder $\mathcal{E}$ and a decoder $\mathcal{D}$, which can achieve the conversion between the pixel space and the latent space. They then train a diffusion model on that latent space with the same objective in Equation~\eqref{equ:simple}. In this work, we will exploit how to fully use the learned high-level knowledge of the pre-trained text-to-image diffusion model in the downstream visual tasks.

\subsection{Prompting Text-to-Image Diffusion Model}
A pre-trained diffusion model contains sufficient information to sample from the data distribution since the model $\bm{\epsilon}_\theta$ can be viewed as the learned gradient of data density $\nabla_{\bm{z}_t} \log p(\bm{z}_t|\mathcal{C})$~\cite{batzolis2021conditional}. As for the text-to-image model, we believe that there is enough high-level knowledge due to the weak supervision of the natural language during pre-training. Our goal is to fully exploit the knowledge of a well-trained text-conditioned $\bm{\epsilon}_\theta$ and transfer the learned knowledge to downstream visual perception tasks. A general perception task aims to model the distribution $p(\bm{y}|\bm{x})$, where $\bm{y}$ is the task-specific label and $\bm{x}$ is the input image. Our basic idea is to build a connection between the task-specific label and the natural language, such that the learned semantic information can be efficiently extracted. To achieve this, we first rewrite the prediction model as $p_\phi(\bm{y}|\bm{x}, \mathcal{S})$, where $\mathcal{S}$ is a set containing all the category names of the task. This is reasonable since the label $\bm{y}$ is related to the $\mathcal{S}$ in both shape and semantic meaning. We then implement $p_\phi(\bm{y}|\bm{x}, \mathcal{S})$ as:
\begin{equation}
    p_\phi(\bm{y}|\bm{x}, \mathcal{S}) = p_{\phi_3}(\bm{y}|\mathcal{F})p_{\phi_2}(\mathcal{F}|\bm{x}, \mathcal{C})p_{\phi_1}(\mathcal{C}|\mathcal{S}),
    \label{equ:decompose}
\end{equation}
where $\mathcal{F}$ is a set of feature maps and $\mathcal{C}$ denotes the text features. We now describe each term in Equation~\eqref{equ:decompose} and its instantiation in detail: 

\noindent (1) \underline{$p_{\phi_1}(\mathcal{C}|\mathcal{S})$} is responsible to extract text features from the class names. We use the same CLIP~\cite{radford2021clip} text encoder as the pre-training stage of Stable-Diffusion~\cite{rombach2022high}, and the text inputs are simply defined using a template of ``\texttt{a photo of a [CLS]}''. However, the domain gap is usually witnessed when transferring the text encoder to downstream tasks~\cite{zhou2022coop,gao2021clipadapter}. Inspired by~\cite{gao2021clipadapter}, we use a text adapter implemented as a two-layer MLP to refine the text features obtained by the CLIP text encoder. To sum up, the text features are computed as follows:
\begin{equation}
\begin{split}
 &\mathcal{T} \leftarrow \{\mathrm{template}(s)|s\in\mathcal{S}\}\\
 &\mathcal{C} \leftarrow \CLIPTextEncoder(\mathcal{T})\\
 &\mathcal{C} \leftarrow \mathcal{C} + \gamma \MLP(\mathcal{C}),
\end{split}
\end{equation}
where $\mathcal{T}$ denotes the raw texts generated by applying the prompt template to the set of class names and $\gamma$ is a learnable scale factor that is initialized to be very small (\eg, 1e-4). This design can help us to maximally preserve the pre-trained knowledge of the text encoder, as well as mitigate the domain gap between the pre-training task and the downstream task. Note that different from the usage of CLIP text encoder in~\cite{rombach2022high} where the features of the whole sentence are used, we simply use the feature from the \texttt{[EOS]} token. Therefore, the shape of $\mathcal{C}$ is $|\mathcal{S}|\times C$ where $C$ is the output dimension of the CLIP text encoder. 

\noindent(2) \underline{$p_{\phi_2}(\mathcal{F}|\bm{x}, \mathcal{C})$} aims to extract hierarchical feature maps $\mathcal{F}$ given the input image $\bm{x}$ and the conditioning inputs $\mathcal{C}$. Since $\mathcal{C}$ contains information from the natural language, $p_{\phi_2}$ needs to capture the cross-domain interactions between vision and language. Interestingly, we find the pre-trained text-to-image diffusion model can be a very good initialization of $p_{phi_2}$. Although $\bm{\epsilon}_\theta$ is trained to perform score-matching~\cite{song2021score} according to the training objective, it has already bridged the vision and language domains. In our implementation, we first use the encoder of the VQGAN $\mathcal{E}$ to encode the image into the latent space (\eg, $\bm{z}_0=\mathcal{E}(\bm{x})$) and then feed the latent feature map and the conditioning inputs to the pre-trained $\bm{\epsilon}_\theta$ network. Note that we simply set $t=0$ such that no noise is added to the latent feature map. The hierarchical features $\mathcal{F}$ can also be easily obtained from the last layer of each output block in different resolutions. Typically, the size of the input image is $512\times 512$ and $\mathcal{F}$ contains 4 feature maps, where the $i$-th feature map $F_i$ has the spatial size of $H_i=W_i=2^{i+2}$, $i=1,2,3,4$.

\noindent(3) \underline{$p_{\phi_3}(\bm{y}|\mathcal{F})$} is the prediction head that generates results from the hierarchical feature maps $\mathcal{F}$. We implement $p_{\phi_3}$ as a Semantic FPN~\cite{kirillov2019semanticfpn}, consisting of several convolutional layers and upsampling layers. The prediction head can be designed to be very lightweight since the $\bm{\epsilon}_\theta$ already has enough capacity to perform downstream vision tasks.

The above formulation enables us to decompose the general visual perception tasks such that the role of the pre-trained diffusion model can be better understood. By injecting the task-specific labels $\mathcal{S}$ as the inputs, we implicitly prompt the pre-trained denoising autoencoder to explore the learned semantic knowledge. It is also worth noting that our method is not a diffusion-based framework anymore, because we only use a single UNet as a backbone (see~\Cref{fig:idea} to better understand the differences). 

\subsection{Semantic Guidance via Cross-attention}

Apart from designing proper prompts to implicitly extract high-level knowledge from $\bm{\epsilon}_\theta$ network, we also propose to use the cross-attention map as an explicit semantic guidance. It has been observed in~\cite{hertz2022prompt} that in a well-trained text-to-image diffusion model, the cross-attention map between the feature map and the conditioning text feature enjoys good locality. This nice property motivates us to leverage the cross-attention maps to explicitly facilitate downstream visual perception. The cross-attention operation exists in each of the 4 resolutions of the \epstheta{} network. Therefore, for the $i$-th resolution, we can simply average all the cross-attention maps belonging to the resolution to obtain an averaged map $A_i$. Since the cross-attention maps are computed by using the conditioning inputs $\mathcal{C}$ as the key and value, the averaged attention map has the shape of $A_i\in \mathbb{R}^{|S|\times H_i\times W_i}$. 

The averaged cross-attention map is useful because each channel of it aggregates some semantic information of a certain category. We can then concatenate the averaged cross-attention maps with the original hierarchical feature maps and fed the results to the prediction head, \ie, $F_i\leftarrow [F_i, A_i]$. By default, we do not use the cross-attention maps at the lowest resolution since they are not very accurate (which we will analyze in the experiments). We empirically find that explicit semantic guidance through cross-attention can help our model faster adapt to downstream tasks.

\subsection{Implementation}
We consider three visual perception tasks in this work, including semantic segmentation, referring image segmentation, and depth estimation. Basically, we use a similar architecture for these tasks, as mentioned above. However, there are some differences in minor design, which we will describe as follows. Firstly, the procedure to obtain the conditioning inputs $\mathcal{C}$ slightly differs in different tasks. For semantic segmentation, $\mathcal{S}$ contains the class names in the dataset. For referring image segmentation, we simply use the referring expression (a single sentence) to compute the conditioning inputs $\mathcal{C}$. For depth estimation, we can build the text prompt similarly using the category name of the scene, such as ``kitchen'', ``bathroom'', \etc. Second, the output channels of the task-specific head $p_{\phi_3}(\bm{y}|\mathcal{F})$ are different. Third, the training objective of the three tasks are varied. We use the cross-entropy loss for both semantic segmentation and referring image segmentation, while the Scale-Invariant loss (SI)~\cite{eigen2014silog} is used for depth estimation. 

\section{Experiments}
To verify the effectiveness of our method, we conduct experiments on three visual tasks including  referring image segmentation, semantic segmentation, and depth estimation, covering both high-level and low-level visual perception. We will first present the experimental settings of these tasks and then give our main results. We will also provide detailed ablation studies and analyses of our method.

\subsection{Experiment Setups}
We first provide some common configurations of VPD. For all three downstream tasks, we fix the VQGAN encoder $\mathcal{E}$ and the CLIP text encoder during training. To fully preserve the pre-trained knowledge of the \epstheta{}, we always set the learning rate of \epstheta{} as 1/10 of the base learning rate. We use $\gamma$=1e-4 for the text adapter. The task-specific settings and training details are elaborated as follows.

\label{sec:exp_setup}

\paragrapha{Semantic Segmentation.} The goal of semantic segmentation is to assign pixel-level labels to a given image, which requires a fine-grained high-level understanding of visual content. We evaluate our method on ADE20K~\cite{zhou2017ade}, which consists of 20K images for training and 2K images for validation. Since our method can adapt faster to the downstream tasks, we train our model for 80K iterations using a Semantic FPN~\cite{kirillov2019semanticfpn} by default. We use a global batch size of 16 and set the learning rate as 1e-4. We use the AdamW optimizer with a weight decay of 1e-4 and warming-up iterations of 1500. We adopt the polynomial learning rate scheduler with a power of 0.9 and a minimum learning rate of 1e-6. For the fast schedule (8K iterations), we linear scale the learning rate schedule and set the warming-up iterations to 150. During inference, we use the slide inference with a crop size $512\times 512$ and a stride of $341\times 341$.

\paragrapha{Referring Image Segmentation.} Referring image segmentation aims to find the related object given a natural language expression from an image. We perform experiments on the widely used benchmark RefCOCO~\cite{yu2016refcoco}, RefCOCO+~\cite{yu2016refcoco}, and more challenging G-Ref~\cite{nagaraja2016modeling} datasets with significantly longer expressions. RefCOCO, RefCOCO+, and G-Ref contain around 20K images and 50K annotated objects, with 142,209, 141,564,  and 104,560 annotated expressions respectively. Following common practice, we train our model on the training set and evaluate the validation set. We use the overall intersection-over-union (oIoU) as the metric to compare different methods. As for the decoder head, we follow LAVT~\cite{yang2022lavt} which uses a simple convolution head to fusion the features and generate the semantic prediction. We train our model for 40 epochs with a total batch size of 32. We set the learning rate as 5e-5 and the weight decay as 0.01. As we have multiple expressions on a single image, during the training phase, we randomly choose a language description. In the inference time, we evaluate sequentially and calculate the mean results following common practice. 

\paragrapha{Depth Estimation.} We adopt a widely used benchmark NYUv2~\cite{silberman2012nyuv2} to evaluate our method in depth estimation. NYUv2 contains 24K images for training and 645 images for testing, covering 464 indoor scenes. Following common practice, we report the absolute relative error (REL), root mean squared error (RMSE), and average log10 error between predicted depth $\hat{d}$ and the ground truth depth $d$. We also report the threshold accuracy $\delta_n$ which denotes $\delta_n=\%$ of pixels satisfying $\max (d_i/\hat{d}_i, \hat{d}_i/d_i)<1.25^n$ for $n=1,2,3$. During training, we randomly crop the images to 480$\times$480. We set the learning rate as 5e-4 and train the model for 25 epochs with batch size of 24. The decoder head and other experimental setting is the same as~\cite{xie2022revealing}. We use the flip and sliding windows during testing.

\subsection{Main Results}
In this section, we will provide our main results on three downstream tasks, including semantic segmentation, referring image segmentation, and depth estimation. Apart from training the models using the default schedule, we also perform experiments on a faster scheduler with very few iterations or epochs to show that our method can quickly adapt to downstream visual perception tasks.

\begin{table}[t]
  \centering
  \caption{\textbf{Semantic segmentation with different methods.} We compare our VPD with previous methods including supervised pre-training and self-supervised pre-training. While other methods adopt the UPerNet~\cite{xiao2018unified} segmentation head, we find our VPD can achieve good results with a more lightweight Semantic FPN~\cite{kirillov2019semanticfpn} with smaller crop size and fewer training iterations.}\vspace{3pt}
  \adjustbox{width=\linewidth}{
    \begin{tabu}to 1.1\linewidth{l*{5}{X[c]}}\toprule
    Method & \#Iters & Crop & FLOPs & mIoU$^{\rm ss}$ & mIoU$^{\rm ms}$ \\\bottomrule
    \addlinespace[3.0pt]
    \multicolumn{5}{l}{\textit{supervised pre-training}}\\
    Swin-L~\cite{liu2021swin}  &160K  & 640$^2$ & 647G & 52.1  & 53.5  \\
    ConvNeXt-L~\cite{liu2022convnet} & 160K  & 640$^2$ & 614G & 53.2  & 53.7  \\
    ConvNeXt-XL~\cite{liu2022convnet}  & 160K  &  640$^2$ & 834G & 53.6  & 54.0  \\\bottomrule
    \addlinespace[3.0pt]
    \multicolumn{5}{l}{\textit{self-supervised pre-training}}\\
    MAE-ViT-L/16~\cite{he2022mae} & 126K  &  - & - & 53.6  & - \\\bottomrule
    \addlinespace[3.0pt]
    \multicolumn{5}{l}{\textit{visial-language pre-training}}\\
    CLIP-ViT-B~\cite{rao2021denseclip} & 80K & 640$^2$ & 340G & 50.6 &  51.3 \\\midrule
    \addlinespace[3.0pt]
    \multicolumn{5}{l}{\textit{text-to-image pre-training}}\\
    VPD (Ours)   & 80K & 512$^2$ & 891G & \textbf{53.7}  & \textbf{54.6}  \\\bottomrule
    \end{tabu}%
  }
  \label{tab:seg}%
  \vspace{-10pt}
\end{table}%

\paragrapha{Semantic Segmentation.} Semantic segmentation is a high-level visual perception task that requires per-pixel high-level understanding. We evaluate our VPD on ADE20K~\cite{zhou2017ade} and compare it with previous backbones and pre-training methods. We start by performing experiments on the default training schedule, where we train our model with a Semantic FPN~\cite{kirillov2019semanticfpn} head for 80K iterations. The results can be found in~\Cref{tab:seg}. For fair comparisons, we do not consider complex segmentation heads such as MaskFormer~\cite{cheng2021maskformer}. Instead, we compare the available result with more common segmentation heads like Semantic FPN~\cite{kirillov2019semanticfpn} and UperNet~\cite{xiao2018unified}. The compared methods include self-supervised pre-training (MAE~\cite{he2022mae}) and supervised pre-training (Swin~\cite{liu2021swin} and ConvNeXt~\cite{liu2022convnet}). We report both the single-scale and multi-scale mIoU for all the methods. We show that VPD can achieve 53.7 mIoU$^{\rm ss}$ and 54.6 mIoU$^{\rm ms}$, outperforming pre-trained ConvNeXt-XL~\cite{liu2022convnet} model with comparable computational complexity. Notably, while other methods utilize UPerNet~\cite{xiao2018unified} as the segmentation head and train the model for $>$120K iterations, our model trained for only 80K iterations can achieve better results with a more lightweight Semantic FPN~\cite{kirillov2019semanticfpn} head and 512$\times$512 crop size. We further perform experiments with a faster schedule, where we train our models for only 8K iterations. We report both the single-scale and multi-scale mIoU at 4K/8K iterations, as shown in \Cref{tab:seg_fast}. We use A32 and A64 subscripts to represent cross-attention maps with spatial sizes up to 32 and 64, respectively. For 8K iterations, we find VPD$_{\rm A32}$ surpass all the baseline methods, including those pre-trained on mask image modelling~\cite{he2022mae,peng2022beitv2}, contrastive learning~\cite{caron2021dino} and supervised learning~\cite{liu2022swinv2,liu2022convnet}. For 4K iterations, we show VPD$_{\rm A64}$ can yield better results. The results indicate that VPD has the potential to enhance adaptation to downstream tasks and that incorporating additional semantic guidance from cross-attention maps can expedite its convergence even further.

\begin{table}[t]
  \centering
  \caption{\textbf{Semantic segmentation with fewer training iterations.} We compare the performance of our VPD with previous models with different architectures and different pre-training methods. The performance is measured by the mIoU of single-scale and multi-scale at 4K/8K iterations.}\vspace{3pt}
  \adjustbox{width=\linewidth}{
    \begin{tabu}to 1.15\linewidth{lX[c]X[c]X[c]X[c]}\toprule
    \multicolumn{1}{l}{\multirow{2}[2]{*}{Method~~~~~~~~~~~~~~~~~~~~~~~~}} & \multicolumn{2}{c}{4K Iters} & \multicolumn{2}{c}{8K Iters} \\\cmidrule(lr){2-3}\cmidrule(lr){4-5}
          & mIoU$^{\rm ss}$ & mIoU$^{\rm ms}$ & mIoU$^{\rm ss}$ & mIoU$^{\rm ms}$ \\\midrule
    DINO-ViT-B/8~\cite{caron2021dino} & 32.4 & 31.1      & 40.8 & 39.9 \\
    MAE-ViT-L/16~\cite{he2022mae} &  37.8     &  36.3     &    46.7   & 46.4 \\
    BeiTv2-ViT-L/16~\cite{peng2022beitv2} &  32.1     &  33.6     &  42.9     & 44.7 \\
    SwinV2-L~\cite{liu2022swinv2} & 40.6 &  41.1     & 47.5 & 48.2 \\
    ConvNeXt-XL~\cite{liu2022convnet} & 43.2 &   43.7    & 47.1 &  47.8\\\midrule
    VPD$_{\rm A32}$ & 43.1 & 44.2 & \textbf{48.7} & \textbf{49.5}\\
    VPD$_{\rm A64}$ & \textbf{\textbf{43.9}} & \textbf{44.7} & 47.7 & 49.1\\\bottomrule
    \end{tabu}%
  }
  \label{tab:seg_fast}%
  \vspace{-10pt}
\end{table}%

\paragrapha{Referring Image Segmentation.} Referring image segmentation also involves high-level knowledge of the correspondence between visual content and referring expression texts. We evaluate our VPD on the widely used RefCOCO~\cite{yu2016refcoco}, RefCOCO+~\cite{yu2016refcoco}, and G-Ref~\cite{nagaraja2016gref_umd}. We train our model on the training set and report the overall IoU (oIoU) on the validation set, as shown in~\Cref{tab:refcoco}. Under the default training schedule, our VPD outperforms previous methods by large margins consistently on both two datasets. We also find that when trained for only 1 epoch, our VPD also achieves better overall IoU than previous state-of-the-art LAVT~\cite{yang2022lavt}. We hypothesize that VPD achieves superior performance due to two primary reasons: (1) unlike prior methods that rely on pre-trained language models that lack interactions with the visual modality, our VPD model leverages a pre-trained diffusion model that learned to generate images guided by the text, thereby establishing a natural connection between language and visual domains. (2) the explicit guidance provided by cross-attention maps offers the model an effective starting point for generating accurate segmentation results.

\begin{table}[t]
  \centering
  \caption{\textbf{Referring image segmentation on RefCOCO.} We compare our VPD with previous methods with both the default training schedule and fast schedule (1 epoch) on three benchmark datasets of RefCOCO (RefCOCO, RefCOCO+, and G-Ref). The performance is measured by the overall IoU on the validation set. We show our VPD achieves better performance consistently on all three benchmarks.}\vspace{3pt}
  \adjustbox{width=\linewidth}{
    \begin{tabu}to 1.2\linewidth{Xcccc}\toprule
     Method  & Language  & RefCOCO  & RefCOCO+ & G-Ref \\\midrule
     \multicolumn{4}{l}{\textit{default schedule}}\\\midrule
    MAttNet~\cite{yu2018mattnet} & Bi-LSTM & 56.51 & 46.67 & 47.64 \\
    MCN~\cite{luo2020mcn} & Bi-LSTM & 62.44 & 50.62 & 49.22 \\
    CGAN~\cite{luo2020cgan} & Bi-GRU & 64.86 & 51.03 & 51.01 \\
    LTS~\cite{jing2021lts} & Bi-GRU & 65.43 & 54.21 & 54.40 \\
    VLT~\cite{ding2021vlt} & Bi-GRU & 65.65 & 55.50 & 52.99 \\
    LAVT~\cite{yang2022lavt} & BERT & 72.73 & 62.14 & 61.24 \\\midrule
    VPD & CLIP & \textbf{73.25} & \textbf{62.69} & \textbf{61.96} \\\midrule
     \multicolumn{4}{l}{\textit{fast schedule, 1 epoch}}\\\midrule
    LAVT~\cite{yang2022lavt}& BERT & 52.56 & 29.17 & 40.31 \\\midrule
    VPD & CLIP & \textbf{63.04} & \textbf{40.01} & \textbf{48.11} \\\bottomrule
    \end{tabu}%
    }
  \label{tab:refcoco}%
  \vspace{-10pt}
\end{table}%

\paragrapha{Depth Estimation.} We start by evaluating VPD on depth estimation, a visual perception task that requires low-level per-pixel understanding. We use the popular benchmark NYUv2 and compare VPD with previous methods, as shown in \Cref{tab:depth}. Under the default training schedule, our VPD achieves 0.254 RMSE, establishing the new state-of-the-art. Notably, our method outperforms SwinV2-B/SwinV2-L~\cite{xie2022revealing}, which uses a very strong visual backbone SwinV2~\cite{liu2022swinv2} pre-trained on masked image modeling. Additionally, we verify the fast convergence of VPD by training the model for only one epoch. \Cref{tab:depth} shows that VPD converged much faster than SwinV2-L~\cite{xie2022revealing}: VPD achieves 0.349 RMSE (lower is better) while the RMSE of SwinV2-L~\cite{xie2022revealing} is 0.381. These results further demonstrate that large-scale text-to-image pre-training can be very competitive in downstream visual perception tasks, even compared with the dedicated visual pre-training methods.

\begin{table}[t]
  \centering
  \caption{\textbf{Depth estimation on NYUv2~\cite{silberman2012nyuv2}.} We report the commonly used metrics for depth estimation including RMSE, $\delta_n$, REL and $\log_{10}$ (see \Cref{sec:exp_setup} for details). We show that VPD outperforms previous state-of-the-art methods consistently in all the metrics. We also demonstrate our model converges faster than SwinV2~\cite{xie2022revealing} pre-trained with masked image modeling in the fast training schedule.}\vspace{3pt}
  \adjustbox{width=\linewidth}{
    \begin{tabu}to 1.3\linewidth{Xcccccc}\toprule
    Method & RMSE$\downarrow$  & $\delta_1\uparrow$    & $\delta_2\uparrow$    & $\delta_3\uparrow$    & REL $\downarrow$  & log10 $\downarrow$ \\\midrule
    \multicolumn{7}{l}{\textit{default schedule}}\\
    \midrule
    BTS~\cite{lee2019big}   & 0.392 & 0.885 & 0.978 & 0.995 & 0.110  & 0.047 \\
    AdaBins~\cite{bhat2021adabins} & 0.364 & 0.903 & 0.984 & 0.997 & 0.103 & 0.044 \\
    DPT~\cite{ranftl2021vision}   & 0.357 & 0.904 & 0.988 & 0.998 & 0.110  & 0.045 \\
    P3Depth~\cite{patil2022p3depth} & 0.356 & 0.898 & 0.981 & 0.996 & 0.104 & 0.043 \\
    NeWCRFs~\cite{yuan2022new} & 0.334 & 0.922 & 0.992 & 0.998 & 0.095 & 0.041 \\
    SwinV2-B~\cite{liu2022swinv2}  & 0.303 & 0.938 & 0.992 & 0.998 & 0.086 & 0.037 \\
    SwinV2-L~\cite{liu2022swinv2} & 0.287 & 0.949 & 0.994 & 0.999 & 0.083 & 0.035 \\
    AiT~\cite{ning2023all} & 0.275 & 0.954 & 0.994 & 0.999 & 0.076 & 0.033\\
    ZoeDepth~\cite{bhat2023zoedepth} & 0.270 & 0.955 & 0.995 & 0.999 & 0.075 & 0.032\\
    \midrule
    VPD   & \textbf{0.254} & \textbf{0.964} & \textbf{0.995} & \textbf{0.999} & \textbf{0.069} & \textbf{0.030} \\\bottomrule
    \addlinespace[4.0pt]
    \multicolumn{7}{l}{\textit{fast schedule, 1 epoch}}\\\midrule
    SwinV2-B~\cite{liu2022swinv2}  & 0.462 & 0.819 & 0.975 & 0.995 & 0.133 & 0.059 \\
    SwinV2-L~\cite{liu2022swinv2}  & 0.381 & 0.886 & 0.984 & 0.997 & 0.112 & 0.051 \\\midrule
    VPD   & \textbf{0.349} & \textbf{0.909} & \textbf{0.989} & \textbf{0.998} & \textbf{0.098} &  \textbf{0.043}\\\bottomrule\\
    \end{tabu}%
    }
  \label{tab:depth}%
  \vspace{-22pt}
\end{table}%

\subsection{Analysis}
In this section, we will conduct detailed analyses to further evaluate the effectiveness of each of the components in VPD, as well as demonstrate the scaling potential of it.

\paragrapha{Effectiveness of components of VPD.} We first evaluate the effectiveness of the components presented in \Cref{sec:method}, as is shown in \cref{tab:ablation}. We perform the ablation studies on semantic segmentation, using the same training configurations as~\Cref{tab:seg_fast}. We start from a vanilla usage of the pre-trained \epstheta{} network as our baseline and add the proposed components gradually to verify the contribution of each. For our baseline (the first row), we feed an empty string as the text prompt, such that no effective visual-language interactions are introduced. We find the performance of the baseline is far from satisfactory (\eg, only 46.9 mIoU at 8K iterations). We then apply the text prompts constructed by filling the class names of ADE20K~\cite{zhou2017ade} to the template ``\texttt{a photo of a [CLS]}'', which can improve the mIoU@4K and mIoU@8K by 0.5 and 0.2, respectively. This reveals that a proper text prompt can build the connection between visual and language domains. To further mitigate the domain gap, we employ the text adapter after the CLIP text encoder, which brings significant improvement (42.0$\to$42.9 in mIoU@4K and 47.1$\to$48.0 in mIoU@8K). Finally, we add the cross-attention maps as explicit semantic guidance (the last row of~\Cref{tab:ablation}) and find the mIoU@8K can be further improved by 0.7. These ablation studies clearly demonstrate that our designs in VPD can effectively leverage the pre-trained knowledge of the \epstheta{} via both implicit and explicit guidance.

\paragrapha{Choice of the cross-attention maps.} There are a lot of cross-attention layers in the denoising autoencoder \epstheta{}. Therefore, it becomes a question that which cross-attention maps we should select to provide semantic guidance. Since \epstheta{} is implemented as a UNet~\cite{ronneberger2015unet}, it consists of mainly three groups of blocks including the downsampling blocks, the middle blocks, and the upsampling blocks. Specifically, for an input image of $512\times 512$, the corresponding size of the latent features is $64\times 64$. The downsampling blocks first gradually reduce the spatial size of the feature maps from $64\times 64$ to $8\times 8$, and then feed them to the middle blocks which do not change the spatial size. Finally, the upsampling blocks progressively increase the feature map size back to $64\times 64$ and merge the information via some lateral connections from the downsampling blocks. 

The comparisons of leveraging the cross-attention maps from different locations can be found in the bottom part of~\Cref{tab:ablation}. First, we show that using the cross-attention maps from middle blocks might be harmful to the performance, mainly because the spatial resolution is too low to provide accurate information. Second, we find that the cross-attention maps from both the upsampling blocks and the downsampling blocks can bring considerable improvements and the upsampling blocks seem to be more beneficial to the performance. This is also reasonable because the cross-attention map will become more and more accurate during the forward procedure. Finally, we average both the cross-attention maps from the upsampling and downsampling blocks and demonstrate that they cooperate well and achieve better results in both mIoU@4K and mIoU@8K.

\begin{table}[t]
  \centering
  \caption{\textbf{Ablation studies.} We perform ablations in semantic segmentation on ADE20K~\cite{zhou2017ade} to verify the effectiveness of each of the proposed components in VPD and the influence of the different choices of cross-attention maps. We find that all the proposed components are beneficial and that combining the cross-attention maps in the downsampling blocks and the upsampling blocks yields the best performance.}\vspace{3pt} 
  \adjustbox{width=\linewidth}{
    \begin{tabular}{ccccc}\toprule
    \multicolumn{1}{l}{text prompt} & \multicolumn{1}{l}{text adapter} & \multicolumn{1}{l}{cross attn} & \multicolumn{1}{l}{mIoU 4K} & \multicolumn{1}{l}{mIoU 8K} \\\toprule
    \xmark     & \xmark     & \xmark     &  41.5 & 46.9 \\
    \cmark     & \xmark     & \xmark     & 42.0  & 47.1  \\
    \cmark     & \cmark     & \xmark     & 42.9  & 48.0  \\\midrule
    \cmark     & \cmark     & mid    & 43.0  & 47.8  \\
    \cmark     & \cmark     & down  & 43.2  & 48.2  \\
    \cmark     & \cmark     & up    & 43.2  & 48.5  \\\midrule
    \cmark     & \cmark     & up+down   & 43.1  & \textbf{48.7}  \\\bottomrule
    \end{tabular}%
    }
  \label{tab:ablation}%
  \vspace{-15pt}
\end{table}%

\vspace{-10pt}

\begin{figure}[t]
    \centering
    \includegraphics[width=\linewidth]{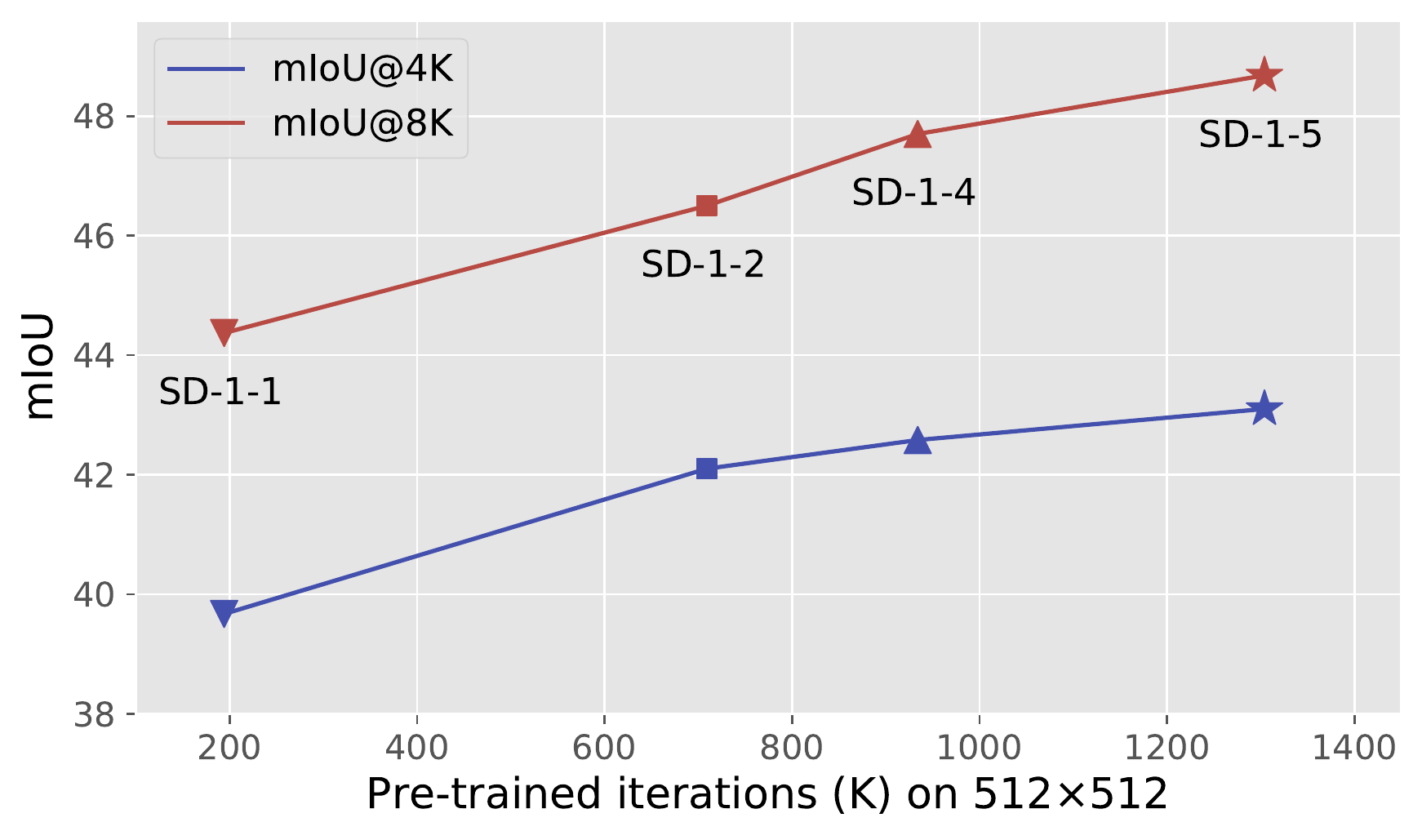}
    \caption{\textbf{Longer pre-training yields better performance on downstream tasks.} We train VPD with different versions of Stable-Diffusion (indicated by \texttt{SD-1-x}) on ADE20K and investigate how the pre-training iteration would affect the performance. The upward trend demonstrates that our VPD can benefit from a stronger text-to-image diffusion model.}
    \label{fig:sd_version}
    \vspace{-12pt}
\end{figure}

\paragraph{Effects of different pre-trained weights.} Since our VPD is built on pre-trained text-to-image diffusion models, it is necessary to investigate how the pre-trained weights would affect the performance of our VPD. In our previous experiments, we have used the released \texttt{1-5} version of the ``Stable-Diffusion'' (\texttt{SD-1-5} for short). Now we compare different releases of ``Stable-Diffusion'' by applying the weights in the semantic segmentation on ADE20K~\cite{zhou2017ade}, and the results are illustrated in~\Cref{fig:sd_version}. The differences between the checkpoints are the pre-training iterations on 512$\times 512$ resolution. We omit the \texttt{SD-1-3} since it is trained for only 30K fewer iterations than \texttt{SD-1-4}. Our results in~\Cref{fig:sd_version} demonstrate a clear trend that more pre-trained iterations of the text-to-image diffusion model will also exhibit better performance on downstream tasks with VPD. It is worth noting that from \texttt{SD-1-1} to \texttt{SD-1-5}, the mIoU@8K is improved by more than 4, which is quite considerable. We hypothesize that this is mainly because longer training can improve the alignment between visual and language, which can be also verified from the CLIP score of the different versions reported by the ``Stable-Diffusion''\footnote{see https://github.com/runwayml/stable-diffusion for details.}. These results also show that the success of our method is based on the learned visual-language knowledge rather than the large capacity of the \epstheta{} network. The upward trend in the graph demonstrates the scaling ability of our VPD, indicating that a stronger pre-trained text-to-image diffusion model can help us achieve better results.

\paragrapha{Limitations. } While our method has shown satisfactory performance, we acknowledge that the computational cost of VPD is currently relatively high. Unlike recognition models that are explicitly designed to balance efficiency and accuracy, generative models prioritize synthesis quality and often lack careful consideration of complexity. Although we have demonstrated the potential of extracting valuable information from a pre-trained text-to-image diffusion model, the high computational costs of \epstheta{} cannot be addressed within our current framework. We believe that further improvements in the complexity-accuracy trade-offs of VPD can be achieved through a more lightweight design of the generative model or a more efficient architecture dedicated to both generative and perception tasks.

\section{Conclusion}
In this paper, we have proposed a new framework called VPD to transfer the high-level knowledge of a pre-trained text-to-image diffusion model to downstream tasks. We have proposed several designs to encourage visual-language alignment and prompt the pre-trained model implicitly and explicitly. Extensive experiments on semantic segmentation, referring image segmentation, and depth estimation have demonstrated that VPD can achieve very competitive performance and exhibits faster convergence compared to methods with various visual pre-training paradigms. We also believe that text-guided generative models other than diffusion models\cite{saharia2022imagen,ramesh2022dalle2,chang2023muse} can also fit in VPD, which we leave to future work. We expect our efforts to shed light on the crucial role of generative text-to-image pre-training in visual perception and make a step towards the unification of visual generation and perception tasks.

{\small
\bibliographystyle{ieee_fullname}
\bibliography{ref}
}

\end{document}